\documentclass[10pt,a4paper,twoside]{article}
\makeatletter\if@twocolumn\PassOptionsToPackage{switch}{lineno}\else\fi\makeatother

\usepackage{amsfonts,amssymb,amsbsy,latexsym,amsmath,tabulary,graphicx,times,xcolor}
\usepackage[T1]{fontenc}
\usepackage{url,multirow,morefloats,floatflt,cancel,tfrupee,multicol}
\makeatletter

\AtBeginDocument{\@ifpackageloaded{textcomp}{}{\usepackage{textcomp}}}
\makeatother
\usepackage{colortbl}
\usepackage{xcolor}
\usepackage{pifont}
\usepackage[nointegrals]{wasysym}
\usepackage[superscript,biblabel]{cite}
\makeatletter

\@ifundefined{etal}{\def\etal{\textit{et~al}}}{}

\usepackage[hmargin=1.8cm,vmargin=2.2cm]{geometry}
\makeatletter
\def\author#1{\gdef\@author{\hskip-\dimexpr(\tabcolsep)\hskip1pt\parbox{\dimexpr\textwidth-1pt}{\centering #1}}}

\let\@articletype\@empty \def\articletype#1{\gdef\@articletype{{\fontsize{14}{16}\selectfont #1}}}

\usepackage{fancyhdr}

\fancypagestyle{headings}{\fancyhf{}\fancyhead[RE]{\MakeTextUppercase{\textbf{\RunningAuthor}}}\fancyhead[LE]{\textbf{\thepage}}\fancyhead[LO]{\MakeTextUppercase{\textbf{\RunningHead}}}\fancyhead[RO]{\textbf{\thepage}}}
\fancypagestyle{plain}{\fancyhf{}\fancyfoot[C]{\textbf{\thepage}}}

\AtBeginDocument{\usepackage{lastpage}}
\def\title#1{%
  \gdef\@title{%
    \ifx\@articletype\@empty\else\@articletype~\\\fi%
     #1}%
}

\usepackage{abstract}
\def\abstractname{\textbf{Abstract}}
\renewenvironment{onecolabstract}
{\vspace*{-.4pc}\trivlist\item[]\leftskip1pt\noindent\selectfont\hfill\abstractname\hfill\mbox{\null}\par\ignorespaces}{\endtrivlist}

\def\NormalBaseline{\def\baselinestretch{1.1}}

\usepackage{textcase}
\usepackage[explicit]{titlesec}
\titleformat{\section}[block]{\NormalBaseline\sffamily\boldmath\bfseries}
{\thesection.}
{6pt}
{#1}
[]
\titleformat{\subsection}[hang]{\NormalBaseline\sffamily\filright\itshape}
{\thesubsection.}
{6pt}
{#1}
[]
\titleformat{\subsubsection}[runin]{\NormalBaseline\sffamily\filright\itshape}
{\hspace{16pt}\thesubsubsection}
{6pt}
{#1}
[]
\titleformat{\paragraph}[runin]{\NormalBaseline\sffamily}
{\theparagraph}
{6pt}
{#1}
[]
\titleformat{\subparagraph}[runin]{\NormalBaseline\sffamily}
{\thesubparagraph}
{6pt}
{#1}
[]

\titlespacing{\section}{0pt}{1.5\baselineskip}{.2\baselineskip}  
\titlespacing{\subsection}{0pt}{1.5\baselineskip}{.2\baselineskip}  
\titlespacing{\subsubsection}{0pt}{1.5\baselineskip}{.2\baselineskip}  
\titlespacing{\paragraph}{0pt}{.5\baselineskip}{10pt}  
\titlespacing{\subparagraph}{0pt}{.5\baselineskip}{10pt}

\usepackage{caption}
\DeclareCaptionLabelFormat{figlabel}{Figure #2}
\date{}
\makeatother

\usepackage{amsmath}
\usepackage{cancel}

\usepackage{float}

\usepackage{xcolor}

\usepackage{color}

\usepackage{cite}
\usepackage[numbers,super]{natbib}

\usepackage{ragged2e}
\makeatletter
\renewenvironment{onecolabstract}{%
  \if@twocolumn
    \section*{\abstractname}%
  \else
    \fontsize{10}{12}\selectfont
    \raggedright 
    \sffamily\bfseries\abstractname
  \fi
  \begin{justify} 
}{%
  \end{justify}
  \vspace{1em} 
}
\makeatother



\makeatletter
\def\title#1{%
  \gdef\@title{%
    \raggedright 
    \ifx\@articletype\@empty\else\@articletype~\\\fi%
    #1%
  }%
}

\usepackage{authblk}

\begin{document}

\title{\textsf{Stiffness Change for Reconfiguration of \\Inflated Beam Robots}}
\def\RunningHead{
\textsf{Stiffness Change for Reconfigurable Robots}
}

\def\RunningAuthor{\textsf{Do \etal}}
\author{\textsf{Brian H. Do, Shuai Wu, Ruike Renee Zhao, and Allison M. Okamura}
\thanks{This work was conducted at the Department of Mechanical Engineering, Stanford University, Stanford, CA 94305 USA.} 
}

\maketitle

{\begin{onecolabstract}
\normalfont
\justifying
Active control of the shape of soft robots is challenging. Despite having an infinite number of \textit{passive} degrees of freedom~(DOFs), soft robots typically only have a few \textit{actively controllable} DOFs, limited by the number of degrees of actuation~(DOAs). The complexity of actuators restricts the number of DOAs that can be incorporated into soft robots. Active shape control is further complicated by the buckling of soft robots under compressive forces; this is particularly challenging for compliant continuum robots due to their long aspect ratios. In this work, we show how variable stiffness can enable shape control of soft robots by addressing these challenges. Dynamically changing the stiffness of sections along a compliant continuum robot can selectively ``activate" discrete joints. By changing which joints are activated, the output of a single actuator can be reconfigured to actively control many different joints, thus decoupling the number of controllable DOFs from the number of DOAs. We demonstrate embedded positive pressure layer jamming as a simple method for stiffness change in inflated beam robots, its compatibility with growing robots, and its use as an ``activating" technology. We experimentally characterize the stiffness change in a growing inflated beam robot and present finite element models which serve as guides for robot design and fabrication. We fabricate a multi-segment everting inflated beam robot and demonstrate how stiffness change is compatible with growth through tip eversion, enables an increase in workspace, and achieves new actuation patterns not possible without stiffening.

\def\keywordstitle{Keywords}
\end{onecolabstract}}


\begin{multicols}{2}

\section{Introduction}

The compliance of soft robots gives them an infinite number of passive degrees of freedom (DOFs). This enables soft robots to adapt to environmental uncertainty even with simple mechanical design, a feature referred to as embodied intelligence.~\cite{rus2015design} These soft robots can be low inertia and inherently safer for interaction tasks with people and objects. They offer the potential to expand the capabilities of robots from the traditional manipulation, inspection, and navigation tasks where rigid robots dominate to new tasks such as physical human-robot interaction.

Compliance also makes active control of soft robots challenging. Only a few of a soft robot's DOFs can be controlled by its actuators. This limits the ability of soft robots to vary their morphology on-demand. Using more actuators can increase the number of actively controllable DOFs but also increases the complexity, weight, and cost of soft robots. Ultimately, this approach can be difficult to scale, with the fabrication and control complexity being limiting factors.

The traditional paradigm in robotics has been to achieve high dexterity through the use of many actuators, each controlling one specific DOF. For example, in the field of robot manipulation, many fully-actuated hand designs exist with an actuator for each finger joint.~\cite{robothandreview} In the field of inflated beam robots, past inflatable robotic arms have relied either on actuators located at the joints~\cite{koren1991inflatable} or cables running from each joint to a base.~\cite{voisembert2013design, perrot2010long} However, this often results in expensive, complicated, or bulky designs which require complex sensing and control.

Designing robots to be underactuated, such as by using passive joints, can reduce the number of actuators used. However, with underactuation, there is a trade-off between the embodied intelligence of robots and their controllability. Multiple DOFs are now coupled to a given actuator, and this mapping between actuators and the set of DOFs these actuators are coupled to cannot be changed after the design stage. 

Stiffness can be used to modulate actuator outputs to yield more complex output shapes.~\cite{shah2021shape, narang2018mechanically} While a variety of approaches have been proposed for stiffness control in soft robots, these focus on binary stiffness change from low to high stiffness states to \textit{maintain} their shape.~\cite{manti2016stiffening} In contrast, biological organisms utilize stiffness to \textit{change} their shape by adapting their morphology. For example, octopuses create virtual joints in their arms by stiffening them during reaching movements,~\cite{yekutieli2005octopus} and elephants actively stiffen their trunks during grasping to create a virtual joint whose location depends on the size of the object they are grasping.~\cite{yekutieli2005octopus}

\begin{figure*}[ht]
    \centering
    \includegraphics[width=\textwidth]{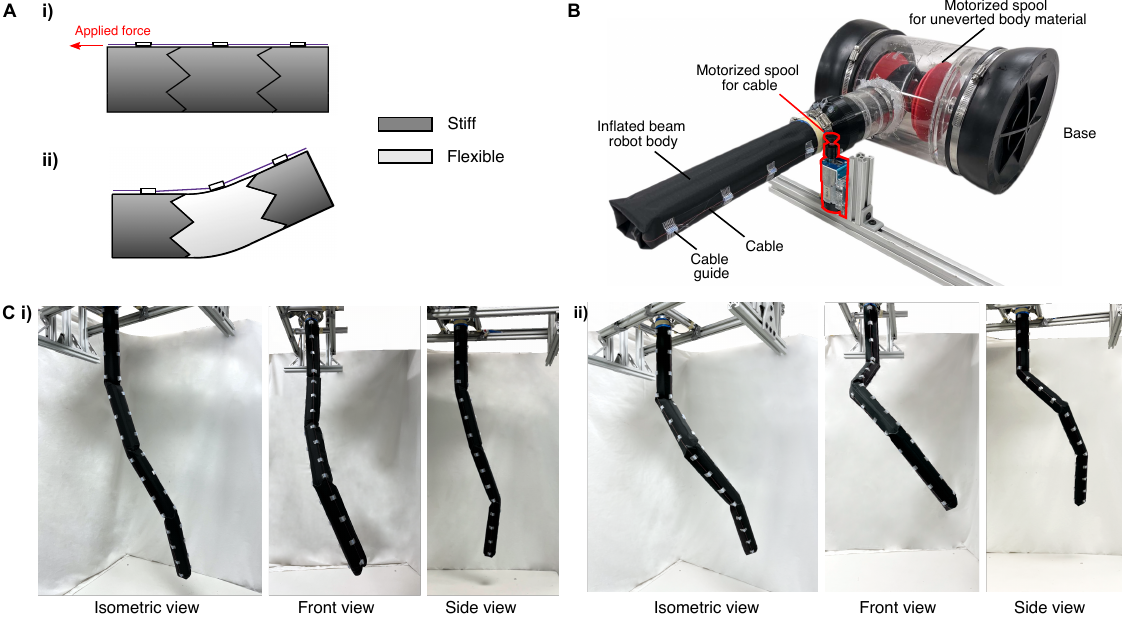}
    \caption{Overview of stiffening vine robot. A)~Schematic of a continuum robot discretized into sections which can be stiff or flexible. In our everting inflated beam robot, stiffening is from layer jamming. i)~Jammed sections are able to resist higher applied forces. ii)~By unjamming a section and then pulling a cable, the inflated beam can be locally buckled at the interface between stiff and flexible sections, causing the overall beam to bend. This bend acts like a revolute joint. B)~Standalone variable stiffness vine robot system. The inflated beam body grows from a stationary base which stores the uneverted body material. The robot is actuated by cables connected to motorized spools at the base. C)~Vine robot in free space actuated in two different configurations.}
    \label{fig:Overview}
\end{figure*}

In this work, we explore how stiffness change can enable reconfiguration of soft inflated beam robots. Inflated beam robots are characterized by their use of internal gas pressure as the supporting element.~\cite{comer1963deflections} Like other continuum robots, inflated beam robots are attractive for human-robot interaction due to their low inertia and have long, thin-aspect ratios.~\cite{sanan2009robots} We focus on the class of inflated beam robots capable of pressure-driven tip eversion, often referred to as ``vine" robots. These everting robots use a flexible but inextensible tube as a pneumatic backbone and can ``grow" by pneumatically everting material at their tip, enabling these robots to be stored compactly and achieve high extension ratios.~\cite{HawkesScienceRobotics2017, mishima2003development, tsukagoshi2011tip} These robots have previously been investigated for manipulation,~\cite{StroppaICRA2020} navigation,~\cite{greer2020robust} and inspection. 

Building on our initial explorations,~\cite{exarchos2022icra, do2020icra} we show how stiffness change can be leveraged for active reconfiguration of soft robots through the design, modeling, and fabrication of an everting inflated beam robot with distributed variable stiffness. Stiffness is used to modify the output of actuators, enabling a single actuator to selectively control multiple independent DOFs and which joint(s) are bent. Transforming the output of a few complex actuators can enable greater adaptive morphing of high-DOF robots. We achieve stiffness change through layer jamming, including positive pressure layer jamming, and characterize this stiffness change both experimentally and in simulation. Finally, we present demonstrations and investigate how stiffness change can be used to expand robot workspace and demonstrate complex shape change in free space.


\section{Materials and Methods} \label{Sec:Methods}

\subsection{Design Concept} \label{Sec:Concept}

\begin{figure*}[!htbp]
    \centering
    \includegraphics[width=\textwidth]{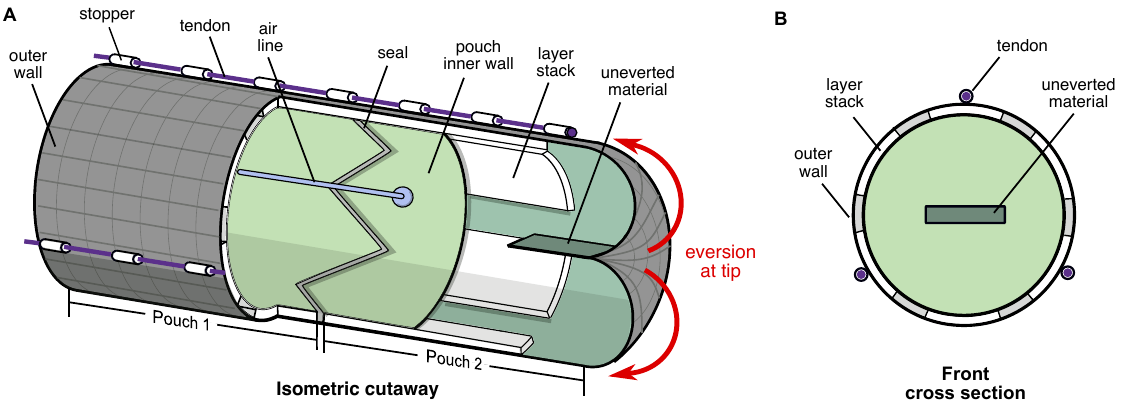}
    \caption{Labeled views of the distributed discrete stiffness inflated beam robot from A)~an isometric cutaway perspective and B)~from a front cross section.}
    \label{fig:Breakdown}
\end{figure*}

Fig.~\ref{fig:Overview}A illustrates how stiffness change enables reconfiguration in inflated beam robots. If we discretize the inflated beam robot into sections, each of which can be independently stiffened, that patterned stiffness can enable bending at specified points. An inherent property of inflated beam robots is that, given a sufficient load, they will undergo either axial or transverse buckling.~\cite{coad2020retraction} The force required to induce buckling in an inflated beam is a function of the material properties of its membrane. In the case where the skin of the robot is sufficiently stiff, the robot will remain straight, as Fig.~\ref{fig:Overview}A~i) shows. However, if we reduce the membrane stiffness of one of the sections, we can induce bending, as Fig.~\ref{fig:Overview}A~ii) shows. In this post-buckled state, a local kink is formed at the interface between stiffened and non-stiffened sections, enabling bending.~\cite{liu2016wrinkling} Bends at these points behave analogously to rotations of revolute joints. Thus, stiffness change enables the creation of specific ``revolute joints" in an otherwise continuum structure. 

Joints can be formed at specified locations by selectively stiffening all sections where bending is not desired. Afterwards, these bends can be locked in place by stiffening the bent sections. This process of expressing which joints are active through stiffening and bending, and then locking bends in place, can be repeated to produce a desired final shape. Furthermore, multiple bends can be formed simultaneously, with the sequence and magnitude of bending tuned by the stiffness of each section. By making one section less stiff than another section, we can induce more bending in the former. The robot configuration can be reset by unstiffening all sections.

Conventionally, each DOF requires its own actuator for independent control.  Patterning stiffness enables a desired DOF to be ``activated" for an actuator to act on. Here, each actuator is mapped to all joints, but a joint can be selectively activated by stiffening. Although stiffening does no work by itself, it modifies actuator outputs to achieve desired work. Stiffness thus modifies the location and magnitude of actuator output. This actuator-activator paradigm reduces the number of complex actuators required while retaining the dexterity of the high-DOF robot system.


\subsection{Body Design for Activation} \label{Sec:BodyDesign}
\subsubsection{Principles}
Stiffness change for everting inflated beam robots is subject to a unique constraint: eversion. Any method for stiffness change must be compatible with pressure-driven eversion. During eversion, material undergoes very high curvature at the tip of the robot. Stiffening mechanisms must be able to accommodate this high curvature. Furthermore, as a result of eversion, robots may grow to meters in length, so any stiffness change method should scale to long lengths. 

Roboticists have investigated several approaches for stiffness change, such as by inducing a physical change to a material's state, such as with low melting point alloys~\cite{Firouzeh2017, McEvoy2016} or phase changing alloys.~\cite{Alambeigi2016} One challenge of these approaches is their slow transition between states.

In this work, we use jamming for stiffening. Jamming is a structural stiffening method in which a pressure difference between particles,~\cite{amend2012positive, steltz2010jamming} layers,~\cite{Ou2014} or fibers~\cite{brancadoro2020fiber} results in the bulk structure behaving cohesively, resulting in a state transition from ``flexible" (unjammed) to ``stiff" (jammed); this phase transition can occur within seconds. Stiffness can be tuned by varying the pressure difference, and once the stiffness is set, no energy is required to maintain it.

The particular jamming approach we use is layer jamming. Layer jamming has a thin form factor, a very high stiffness change, and lower volume compared to other jamming techniques. This enables compatibility with tip eversion. Traditional jamming has relied on a vacuum source to generate negative pressure to produce the requisite pressure difference. We also achieve layer jamming without a vacuum source by utilizing the pressure difference between the inflated robot internal pressure and atmospheric pressure.~\cite{do2020icra, do2022all}

\subsubsection{Design Overview}

Fig.~\ref{fig:Breakdown} shows a schematic of our everting inflated beam robot with distributed variable stiffness. The main body is a tube of material which is initially inverted. Uneverted material is pulled through the inside of the robot before being everted at the tip. The robot skin is composed of two walls, and the space between these inner and outer walls forms pouches where the jamming layers are located. The robot is discretized into sections, each consisting of a pouch running circumferentially around the robot and containing 6 stacks of paper, with 15 layers each; the layer stacks are described in more detail in Sec.~\ref{Fab}. We actuate the robot using tendons and stoppers which run along the length of the robot. 

\subsubsection{Fabrication} \label{Fab}

\begin{figure*}
    \centering
    \includegraphics[width=\textwidth]{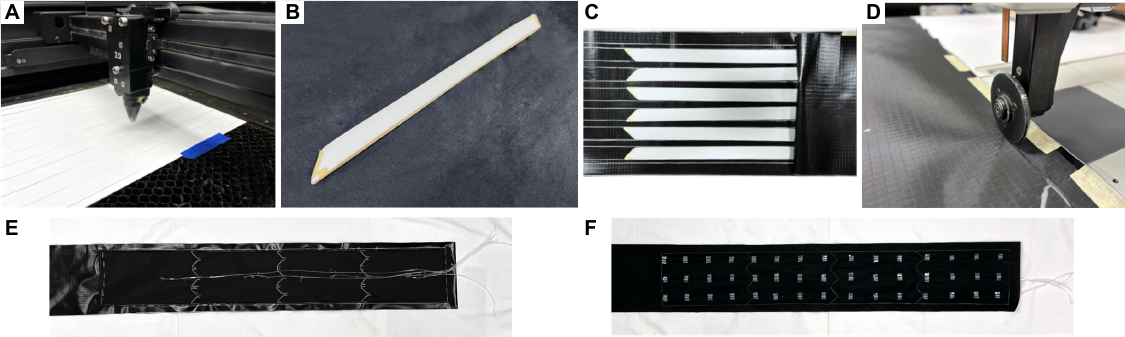}
    \caption{Fabrication process for the stiffening vine robot. A)~Layers are first cut from copy office paper using a laser cutter. B)~These are assembled into layer stacks. C)~Six layer stacks are secured onto TPU-coated 70D-ripstop nylon fabric. D)~Pouches are formed by sealing the fabric using an ultrasonic welder. E)~Overhead view of inner vine skin before sealing, with tubes embedded into each pouch. F)~Overhead view of outer vine skin before sealing, with stoppers aligned along three lines.}
    \label{fig:Fabrication}
\end{figure*}

Fig.~\ref{fig:Fabrication} shows the step-by-step fabrication process for the robot. 

The body of the inflated beam robot is a 55~mm diameter tube fabricated from flat sheets of 70 denier ripstop nylon (Quest Outfitters, Sarasota, FL) with an airtight thermoplastic polyurethane (TPU) coating.

Layers are 0.1~mm thick sheets of copy paper (Amazon Basics Multipurpose Copy Paper) cut using a laser cutter (Universal Laser Systems, Inc., Scottsdale, AZ) into thin strips. Paper was chosen as the layer material due to its low thickness, consistent surface frictional properties, and low cost. 

To facilitate eversion of the layers, the layer width was set at 18~mm, leaving a 10.8~mm gap between adjacent layer stacks. As the robot is growing, uneverted material must transform from a flat, folded state into a curved surface as it forms the beam wall. This necessitates a change in the Gaussian curvature of the material. From Gauss's \textit{Theorema egregium}, this would not be possible for a single inextensible sheet without wrinkling or creasing.~\cite{vanRees2017pnas} The gaps between layer stacks allow the fabric to crease instead of the layers; the robot can thus evert without distorting the layers. The lengths of strips is set by the desired length of each section.
 
Layer stacks are assembled from 15 layers and secured to the fabric using cyanoacrylate glue. A second sheet of ripstop nylon was then overlaid onto the layer stacks to form the pouch inner wall. A 3.175~mm diameter plastic tube was inserted through the inner wall for each pouch and secured in place using hot glue. Afterwards, pouches were sealed using an ultrasonic welder (Vetron Typical GmbH, Kaiserslautern, Germany). 

In previous work, we demonstrated a reconfigurable inflated beam robot with embedded 3D printed bistate valves.~\cite{do2020icra} While embedding passive valves simplifies fabrication for robots with many sections, for shorter robots, the benefits of direct pouch pressure control, such as reduced response time, can outweigh its costs. 

The tube is formed by folding the fabric onto its itself lengthwise and sealing along the entire length using the ultrasonic welder. The end of the tube is also sealed. 

A bend can be created in an inflated beam robot by applying a compressive force at the tip or through distributed loading along a side. We use tendons and stoppers to induce distributed shortening.~\cite{blumenschein2021geometric} 1~cm long, 5~mm diameter polytetrafluoroethylene (PTFE) stoppers were secured using double-sided tape onto the beam outer wall with 6~cm spacing between adjacent stoppers. Spectra cables were then inserted through the stoppers. One end of the cables is secured to the tube end and the other is attached to the motorized spools. The tube is then inverted, with one end attached to the opening in the base and the other end wrapped around a spool for storage.

\subsection{Joint Design for Actuation}

The interface between adjacent sections is a triangular wave to prevent premature beam buckling. There are three triangles around the beam circumference with each triangle composed of two parallelogram-shaped paper layer stacks. 

The triangular wave interface prevents premature beam buckling by inhibiting the formation of wrinkles in a circular arc around the beam. When inflated beams are exposed to a bending moment, behavior is first defined by global beam bending. As the applied load is increased, local wrinkles form and begin to propagate around its circumference, eventually reaching a failure point at which point, a sharp local kink forms.~\cite{liu2016wrinkling} If the interface between adjacent sections was straight, then these wrinkles could form in the gap between layer stacks without being impeded. The triangular wave interface ensures that wrinkling must involve the jamming skin. Adjacent sections are sealed as close together as possible to minimize the area not covered by the jamming skin. 

The robot bends at the interfaces between sections, and thus, the interfaces can act as temporary revolute joints when actuated by the tendons and stoppers. The tendons provide a force to the robot tip, and each is actuated by a motorized spool located at the robot base. Pulling on a tendon shortens that side of the robot, causing the robot to bend in that direction. Stoppers along the robot's length route the tendons axially. Three tendons are arranged radially around the center of the cross section, 120$^\circ$ from each other, enabling bending in 3-D free space.

\subsection{Robot System}
Fig.~\ref{fig:Overview}B shows the full robot system. The end of the robot body, described in Sec.~\ref{Sec:BodyDesign}, is attached to an opening in a rigid pressure vessel. The body is then inverted and stored on a spool inside the pressure vessel. This forms the robot base. The robot body grows when the base is pressurized. Motorized spools are secured at the front of the robot for actuation.

\begin{figure*}
    \centering
    \includegraphics[width=0.75\columnwidth]{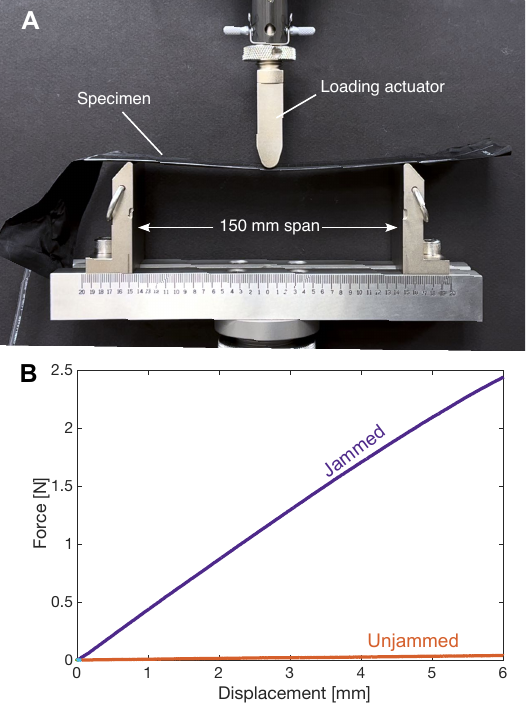}
    \caption{Materials characterization. A)~Three-point bending test setup. B)~Force-displacement curves for stack of 15 layers. For the jammed and unjammed conditions, force curves remain consistent across displacements tested.}
    \label{fig:MaterialCharacterization}
\end{figure*}

\section{Stiffness Characterization} 

The bending behavior of an inflated beam robot depends on its stiffness. We characterized the material stiffness, beam stiffness, and joint stiffness using empirical tests and finite element analysis (FEA) for our inflated beam robot.


\subsection{Material Stiffness}

We obtained material parameters for FEA by conducting three-point bending tests using an Instron universal test machine (Model 3344, Instron, Norwood, MA, USA) on an individual stack of 15 layers of copy paper (Fig.~\ref{fig:MaterialCharacterization}A). The layers with a width \textit{b} of 35 mm were sealed inside the same TPU-coated ripstop nylon described in Sec.~\ref{Fab}, resulting in a total thickness \textit{d} of 1.82 mm.  A beam span \textit{L} of 150 mm was used. The beam, unjammed or jammed
at vacuum pressure, was displaced to 6 mm at a rate of 24 mm/min. Fig.~\ref{fig:MaterialCharacterization}B shows the force-displacement curves of two tests. The flexural modulus is calculated by \(\textit{E} = (\textit{L}^3m)/(4bd^3)\), where \textit{m} is the slope from the initial portion (displacement $<$ 0.5 mm) of the force-displacement curve. The effective material stiffness from the flexural modulus of the unjammed and jammed states are calculated to be 263.2~kPa and 3038.6~kPa respectively.

\subsection{Beam Stiffness Change due to Jamming}

\begin{figure*}
    \centering
    \includegraphics[width=0.9\columnwidth]{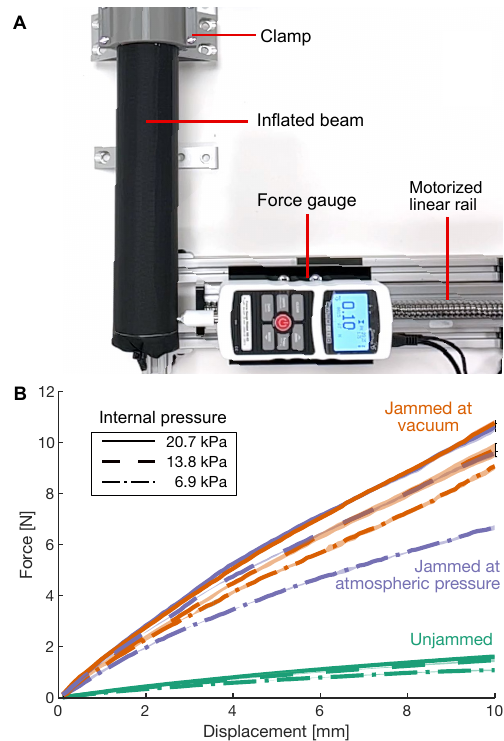}
    \caption{Jamming skins allow beam stiffness to be tuned. A)~Test setup for beam transverse loading using a force gauge on a motorized linear rail. B)~Beam stiffness increases modestly with beam internal pressure and significantly with jamming. Shaded regions represent 1 standard deviation. Solid colored lines represent the mean.}
    \label{fig:SingleSegment}
\end{figure*}

\begin{figure*}[htbp]
    \centering
    \includegraphics[width=\textwidth]{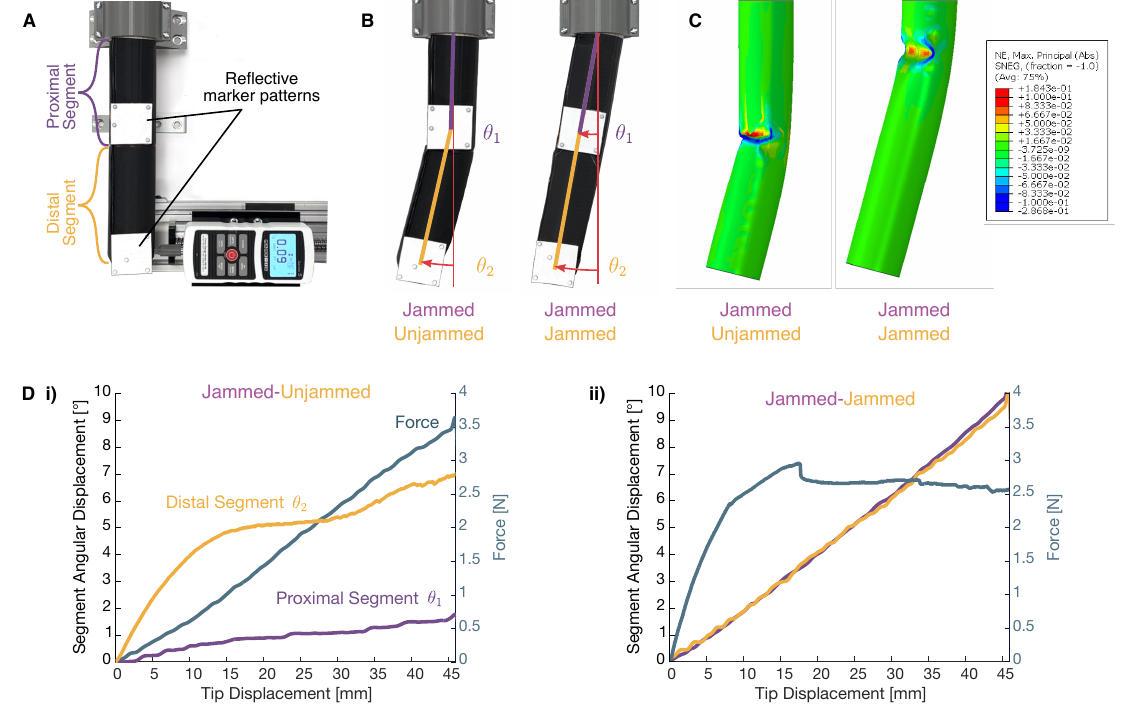}
    \caption{Bending tests with a two-segment stiffening inflated beam robot. A)~Test setup for beam transverse loading. Retroreflective markers are placed at the end of each segment to allow measurement of position and orientation using a motion capture system (not pictured). B)~i)~When the proximal segment is jammed while the distal segment is unjammed, the boundary between the two acts as a pivot. ii)~Jamming both sections results in the entire beam behaving cohesively. C)~FEA depicts the same pattern of bending. Here, strain is shown. D)~When the distal segment is unjammed and the proximal is jammed, the distal has a larger angular displacement. When both are jammed, the two have the same angular-displacement curves. In i) applied transverse force at the tip increases monotonically due to the restoring torque from the bent beam. In ii) force increases linearly then plateaus due to the formation of a crease in the robot's proximal section.}
    \label{fig:TwoSegment}
\end{figure*}

To characterize the effect of jamming skins on the buckling of inflated beam robots, we tested the bending stiffness of a 55~mm diameter, 250~mm long inflated beam with a single jamming section of 6 layer stacks, each composed of 15 layers of copy paper, running along the beam's length. 

Fig.~\ref{fig:SingleSegment}A shows the test setup used to measure the bulk beam stiffness. A clamp secured one end of the inflated beam. A 3D-printed attachment on a Mark-10 Series 5 force gauge applied a transverse force to the other end of the beam. The force gauge was mounted on a motorized linear rail traveling at 0.2~mm/s for a tip deflection of 10~mm. We measured force-displacement curves for three sets of beam internal pressures --  6.9, 13.8, and 20.7~kPa (1, 2, and 3~psi) gauge -- and three sets of pouch pressure differences -- jammed at vacuum, jammed at atmospheric pressure (pouch pressure at atmospheric pressure), and unjammed (pouch pressure at internal pressure). Therefore, for example, for a 6.9~kPa internal pressure, jamming at atmospheric pressure corresponds to a $\Delta P = 6.9$~kPa with the pouch pressure whereas for a 20.7~kPa internal pressure, it corresponds to a $\Delta P = 20.7$~kPa. Five trials were recorded for each of the nine conditions.

Fig.~\ref{fig:SingleSegment}B shows the force-displacement curves. Across conditions, increasing the internal pressure increased the beam stiffness, with diminishing returns on stiffness increase due to the layers providing most of the stiffness. For the unjammed case, increasing internal pressure produces relatively modest stiffness change. In contrast, jamming the layers produces a large increase in stiffness. For a 6.9~kPa internal pressure, the unjammed beam requires 1.07~N to displace 10~mm while the beam jammed at atmospheric pressure requires 6.68~N -- a 624\% increase. For a 20.7~kPa internal pressure, the force increases 663\%. Jamming enables stiffness change otherwise only possible with very high internal pressure. 

There is also a diminishing return to the stiffness change with an increase in the pouch pressure difference. For internal pressures of 13.8~kPa and 20.7~kPa, the mean percent difference between jamming at atmospheric pressure and at vacuum are 4.7\% and 2.1\%, respectively. This is possibly due to the bulk stiffness of the jammed layers being the limiting factor rather than the layers delaminating due to insufficient friction force. The comparable performance indicates that positive pressure jamming is an effective means for varying stiffness and that a negative pressure source is not required.


\subsection{Joint Stiffness}

Fig.~\ref{fig:TwoSegment}A shows the experimental setup for the joint stiffness characterization tests. An inflated beam was fabricated with two independently jamming pouches. A transverse load was applied using the same experimental setup as for the single segment tests as shown in Fig.~\ref{fig:SingleSegment}A. Reflective marker patterns were used to measure the orientation of the front and back segments of the inflated beam using an OptiTrack motion capture camera system with Flex 13 cameras. 

By modifying the bending stiffness of the joint, the bending location of the beam is controlled. Fig.~\ref{fig:TwoSegment}B shows how the joint stiffness affects the bending behavior of the beam. When the segment distal to the base is unjammed, a joint forms at the interface between the proximal and distal segments, resulting in the proximal segment remaining undisturbed while the distal segment bends. When both segments are jammed, the joint stiffness increases and the beam bends as a single unit at its base. 

FEA simulations of the two-segment tube with independently tunable stiffness are carried out using ABAQUS 2021 (Dassault Systèmes, France), as shown in Fig.~\ref{fig:TwoSegment}C. Linear elastic models are used with effective Young's moduli of 263.2 kPa and 3038.6 kPa for the unjammed and jammed regions, respectively, and the same Poisson's ratio of 0.45. Consistent with the experimental results, localized buckling occurs between the two segments of the tube under a transverse load when only the distal segment is unjammed. In contrast, when both segments are jammed, the tube shows no obvious relative rotation between the two segments but deforms mainly at the fixed end of the tube, which agrees well with the experimental results.

Fig.~\ref{fig:TwoSegment}D shows an overlay of the body segment angles over time as well as the applied force producing this change from the empirical testing. The difference in joint bending behavior can be seen by comparing the distal segment angle $\theta_1$ and proximal segment angle $\theta_2$ between the unjammed and jammed cases. When the distal segment is unjammed, the tip deflection is primarily due to the rotation of the distal segment whereas when the distal segment is jammed, the entire structure behaves as a single unit, with the distal and proximal angles tracking together.

\section{Demonstrations: Active Shape Change \\and Reconfiguration}
\subsection{Growth and Bending}

\begin{figure*}[htbp]
    \centering
    \includegraphics[width=\columnwidth]{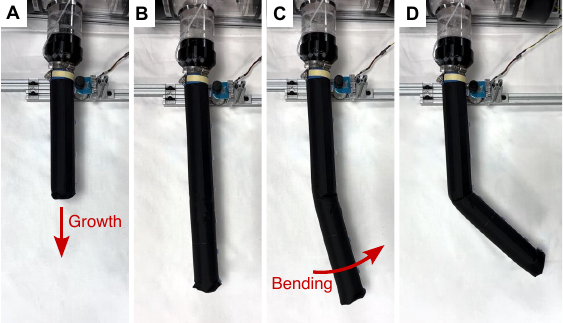}
    \caption{Actuation sequence of a two-segment vine robot. A)~The robot can grow through pressure-driven eversion to a desired length. B)~Once the robot reaches a desired length, a desired stiffening pattern can be induced to select a pivot point and then a tendon can be pulled to initiate bending. C)~Bending occurs about a pivot point. D)~The final bend configuration.}
    \label{fig:GrowThenBend}
\end{figure*}

We can achieve growth, bending, and retraction with our robot system. Initially, the robot can be compactly stored on a spool in an airtight base, from which it can be grown to a specified length. During growth, the pouches are pressurized to the internal body pressure to ensure that the layers remain unjammed, enabling them to evert at the tip.  Following growth, the robot can be bent via tendons into a desired configuration. Fig.~\ref{fig:GrowThenBend}A illustrates this growth and bending sequence for a two-segment vine robot.


\subsection{Workspace}
\begin{figure*}
    \centering
    \includegraphics[width=\textwidth]{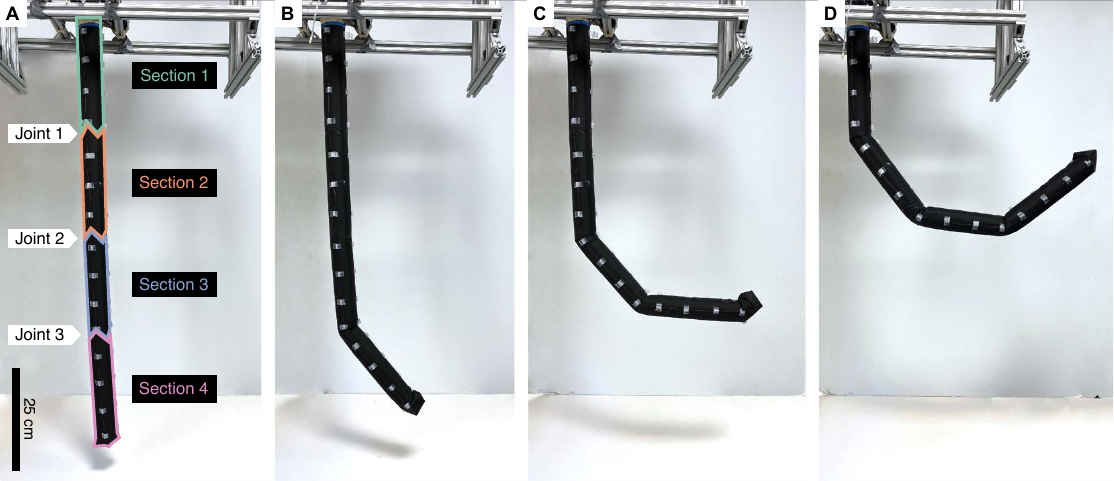}
    \caption{Multibend workspace. Joints can be selectively activated and bent based on the patterned stiffness. A)~Annotated photo showing location of each pouch and joint. Each joint can be selectively activated. B)~Here, the joint between Sections 3 and 4 is bent while the rest of the robot remains rigid and straight. C)~Now, Sections 3 and 4 are stiffened, locking the bend between them in place. The joint between Sections 2 and 3 can now be bent. D)~This process can be repeated until all joints are bent.}
    \label{fig:MultibendWorkspace}
\end{figure*}

For this demonstration, a 1~m long robot consisting of four 0.25~m long sections was fabricated. The robot was suspended from the ceiling and was actuated using three independently actuated tendons running along the length of the robot to enable motion in free space. Fig.~\ref{fig:MultibendWorkspace}A shows the unactuated robot.

The addition of reconfigurable discrete joints expands the workspace of the robot. Previous work using cables to actuate everting inflated beam robots were unable to prevent buckling and had a single revolute joint at the base about which the rest of the robot pivoted.~\cite{StroppaICRA2020} Incorporating variable stiffness sections into our inflated beam robot enables us to exert compressive forces using the cables while avoiding buckling. The reconfigurable joints at the interfaces between sections can be selectively activated, enabling the robot to bend at these joints.

Fig.~\ref{fig:MultibendWorkspace} shows a time series of images showing the sequential forming and bending of temporary revolute joints, with each bend targeting $30^{\circ}$ between adjacent segments. This configuration was generated by sequentially unstiffening and restiffening segments, starting at the most distal segment (Section 4) and moving towards the base. Using a single cable, each joint can be independently actuated from the others due to the stiffness patterning. For example, Fig.~\ref{fig:MultibendWorkspace}B is generated by maximally jamming Sections 1-3, unjamming Section 4, and pulling on one of the tendons. After the desired bend angle is achieved, Section 4 was re-jammed, locking the bend in place during subsequent bends. This process was then repeated for Section 3 and for Section 2.

There are several advantages of this approach compared to other state-of-the-art robots. For example, piecewise constant curvature robots require different actuation inputs for each section. Thus, generating a similar shape to that in Fig.~\ref{fig:MultibendWorkspace}D would require the use of 3 different actuators. While the robot configuration shown in Fig.~\ref{fig:MultibendWorkspace} could be achieved with a similarly designed global constant curvature robot with a single actuator, such a robot would suffer in its precision for targets located at a small angular displacement away from the vertical. By bringing the hinge of the joint close to the end rather than initiating bending at the robot base, we can minimize the effect of actuation errors on end effector position.

\subsection{Multibend Variable Stiffness}

\begin{figure*}
    \centering
    \includegraphics[width=0.8\textwidth]{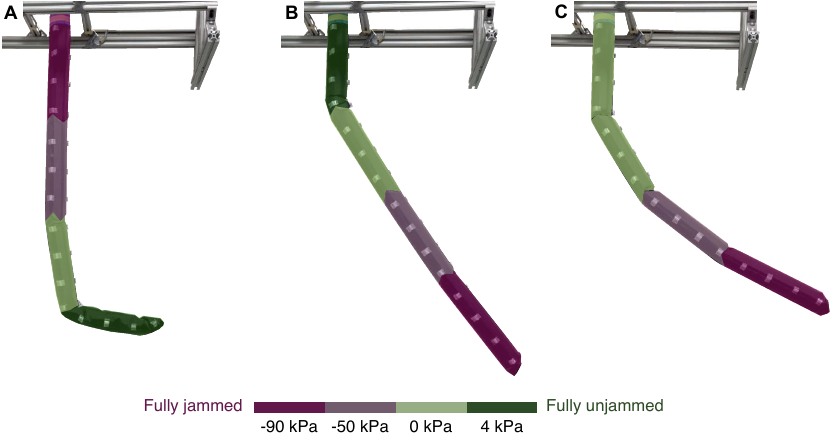}
    \caption{Variable stiffness reconfiguration. The same actuator inputs yield three different output robot configurations, with the location of bends determined by the patterned stiffness. A false color overlay over each robot shows the pressure in the robot skin of each section.}
    \label{fig:MultibendVariableStiffness}
\end{figure*}

Varying the programmed stiffness pattern yields different output configurations from the same actuator input(s). Each joint in the robot acts as a universal revolute joint with a torsional stiffness dependent on the stiffness of its adjacent sections. The relative stiffness of these joints can be tuned by changing the pressures in each section. Given the application of a force at the tip via a tendon, the resulting displacements are determined by the relative joint stiffnesses and relative applied moment. 

Fig.~\ref{fig:MultibendVariableStiffness} shows how three different stiffness patterns result in three different configurations. A false color overlay illustrates the pressure of each section. Fig.~\ref{fig:MultibendVariableStiffness}A depicts the resulting configuration for a stiffness pattern that begins fully jammed at the robot base and ends fully unjammed at the robot tip while Fig.~\ref{fig:MultibendVariableStiffness}B depicts the resulting configuration for the reversed stiffness pattern that begins fully unjammed at the robot base and fully jammed at the robot tip. In Fig.~\ref{fig:MultibendVariableStiffness}A, bending occurs at both Joints 2 and 3 whereas in Fig.~\ref{fig:MultibendVariableStiffness}B, bending occurs at Joint 1 and is negligible at the other joints. Fig.~\ref{fig:MultibendVariableStiffness}C features the same stiffness pattern as Fig.~\ref{fig:MultibendVariableStiffness}B with the exception of Section 1, which is now at 0~kPa, resulting in Section 1 being stiffer in Fig.~\ref{fig:MultibendVariableStiffness}C than in Fig.~\ref{fig:MultibendVariableStiffness}B. As a result of the more similar stiffnesses between joints, Fig.~\ref{fig:MultibendVariableStiffness}C now features much more significant bending at Joint 2 as well as Joint 3.

\subsection{3D Shape Change in Free Space}

Fig.~\ref{fig:Overview}C shows two example robot configurations in free space. We actuate the robot to form shapes in free space using just three tendons which run along the entire length of the robot. Through stiffness change of the robot sections, these three tendons can be used to independently actuate each robot joint. Each joint can be bent by pulling either an individual cable or multiple cables simultaneously and thus function like a 2-DOF universal joint.


\section{Conclusion}
This article presents the design, characterization, and fabrication of growing inflated beam robots which leverage stiffness change to selectively activate dynamically reconfigurable discrete joints. In doing so, decoupling the number of controllable DOFs from DOAs. We characterized the behavior of one- and two-segment stiffening beams experimentally and performed FEA. We also fabricated multi-segment growing inflated beam robots and demonstrated how stiffness change is compatible with pressure-driven eversion, enables a larger robot workspace, yields multibend variable stiffness, and 3D shape change in free space.

The actuation-activation paradigm enabled by the selective activation of joints through stiffness change allows the mapping between actuators and the DOFs they control to be varied. A single actuator can now independently control many DOFs and the same actuator inputs can yield different shape outputs, thereby reducing the number of required actuators and simplifying active shape change. This actuation-activation paradigm could be applied across different types of soft robots to enable shape change.

Our soft robot possesses growth, variable stiffness control, and both continuum links and variable discrete joints. Future work will investigate leveraging these properties for manipulation. By combining the advantages of both traditional rigid and soft manipulators, future systems could enable more collaboration between humans and robots. Additionally, we would like to incorporate sensing to enable closed-loop control. Beyond manipulation, stiffness change across different diameter and length scales could be useful for robots in many applications, ranging from minimally-invasive surgery to reconfigurable structures.

\section*{Acknowledgments}
This work was supported in part by the National Science Foundation Graduate Research Fellowship Program; National Science Foundation grants 2024247 and 2145601; the U.S.\ Department of Energy, National Nuclear Security Administration, Office of Defense Nuclear Nonproliferation Research and Development (DNN R\&D) under subcontract from Lawrence Berkeley National Laboratory; and the United States Federal Bureau of Investigation contract 15F06721C0002306.

\section*{Author Disclosure Statement}
No competing financial interests exist.

\bibliographystyle{vancouver}
\renewcommand{\bibnumfmt}[1]{#1.}
\bibliography{References}

\section*{}

\raggedleft
Address correspondence to: \\
\textit{
Brian Do \\
Department of Mechanical Engineering \\
Stanford University \\
Stanford, CA 94305 \\
USA}

\vspace{1em}
\textit{Email:} brian.do@oregonstate.edu

\end{multicols}
\end{document}